\documentclass[sigconf]{acmart}

\usepackage[ruled,vlined]{algorithm2e}
\usepackage{multirow}
\usepackage{dsfont}

\AtBeginDocument{%
  \providecommand\BibTeX{{%
    \normalfont B\kern-0.5em{\scshape i\kern-0.25em b}\kern-0.8em\TeX}}}

\copyrightyear{2022} 
\acmYear{2022} 
\setcopyright{acmcopyright}\acmConference[KDD '22]{Proceedings of the 28th ACM SIGKDD Conference on Knowledge Discovery and Data Mining}{August 14--18, 2022}{Washington, DC, USA}
\acmBooktitle{Proceedings of the 28th ACM SIGKDD Conference on Knowledge Discovery and Data Mining (KDD '22), August 14--18, 2022, Washington, DC, USA}
\acmPrice{15.00}
\acmDOI{10.1145/3534678.3539181}
\acmISBN{978-1-4503-9385-0/22/08}

\begin{document}

\title{Device-cloud Collaborative Recommendation via Meta Controller}

\author{Jiangchao Yao}
\affiliation{%
\institution{CMIC, Shanghai Jiao Tong University}
\city{Shanghai} 
\country{China}}
\email{Sunarker@sjtu.edu.cn}

\author{Feng Wang}
\affiliation{%
\institution{DAMO Academy, Alibaba Group}
\city{Hangzhou} 
\country{China}}
\email{wf135777@alibaba-inc.com}

\author{Xichen Ding}
\affiliation{%
\institution{Ant Group}
\city{Beijing} 
\country{China}}
\email{xichen.dxc@antgroup.com}

\author{Shaohu Chen}
\affiliation{%
\institution{Ant Group}
\city{Beijing} 
\country{China}}
\email{shaohu.csh@antgroup.com}

\author{Bo Han}
\affiliation{%
\institution{Hong Kong Baptist University}
\city{Hong Kong} 
\country{China}}
\email{bhanml@comp.hkbu.edu.hk}

\author{Jingren Zhou}
\affiliation{%
\institution{DAMO Academy, Alibaba Group}
\city{Hangzhou} 
\country{China}}
\email{jingren.zhou@alibaba-inc.com}

\author{Hongxia Yang}
\affiliation{%
\institution{DAMO Academy, Alibaba Group}
\city{Hangzhou} 
\country{China}}
\email{yang.yhx@alibaba-inc.com}

\renewcommand{\shortauthors}{Jiangchao Yao et al.}
\newcommand{\cloud}{\text{cloud}}
\newcommand{\device}{\text{device}}
\newcommand{\refresh}{\text{re-fresh}}
\newcommand{\base}{\text{base}}
\newcommand{\uplifta}{\text{uplift1}}
\newcommand{\upliftb}{\text{uplift2}}
\newcommand{\seq}{\text{seq}}
\newcommand{\cbc}{\text{cbc}}
\newcommand{\odc}{\text{odc}}
\newcommand{\meta}{\text{meta}}
\newcommand{\bfX}{\mathbf{X}}
\newcommand{\bfY}{\mathbf{Y}}
\newcommand{\bfH}{\mathbf{H}}
\newcommand{\bfS}{\mathbf{S}}
\newcommand{\bfO}{\mathbf{O}}

\begin{abstract}
  On-device machine learning enables the lightweight deployment of recommendation models in local clients, which reduces the burden of the cloud-based recommenders and simultaneously incorporates more real-time user features. Nevertheless, the cloud-based recommendation in the industry is still very important considering its powerful model capacity and the efficient candidate generation from the billion-scale item pool. Previous attempts to integrate the merits of both paradigms mainly resort to a sequential mechanism, which builds the on-device recommender on top of the cloud-based recommendation. However, such a design is inflexible when user interests dramatically change: 
  the on-device model is stuck by the limited item cache while the cloud-based recommendation based on the large item pool do not respond without the new re-fresh feedback.
  To overcome this issue, we propose a meta controller to dynamically manage the collaboration between the on-device recommender and the cloud-based recommender, and introduce a novel efficient sample construction from the causal perspective to solve the dataset absence issue of meta controller. On the basis of the counterfactual samples and the extended training, extensive experiments in the industrial recommendation scenarios show the promise of meta controller in the device-cloud collaboration. 
\end{abstract}



\maketitle

\section{Introduction}
Recommender system~\cite{resnick1997recommender} has been an indispensable component in a range of ubiquitous web services. With the increasing capacity of mobile devices, the recent progress takes on-device modeling~\cite{lin2020mcunet,dhar2021survey} into account for recommendation by leveraging several techniques like network compression~\cite{sun2020generic}, split-deployment~\cite{gong2020edgerec} and efficient learning~\cite{yao2021dccl}. The merits of on-device real-time features and reduced communication cost in this direction then attract a range of works in the perspectives of privacy~\cite{yang2020federated} and collaboration~\cite{sun2020convergence}, and boost the conventional industrial recommendation mechanism.

\begin{figure}[!t]
  \centering
  \includegraphics[width=0.85\linewidth]{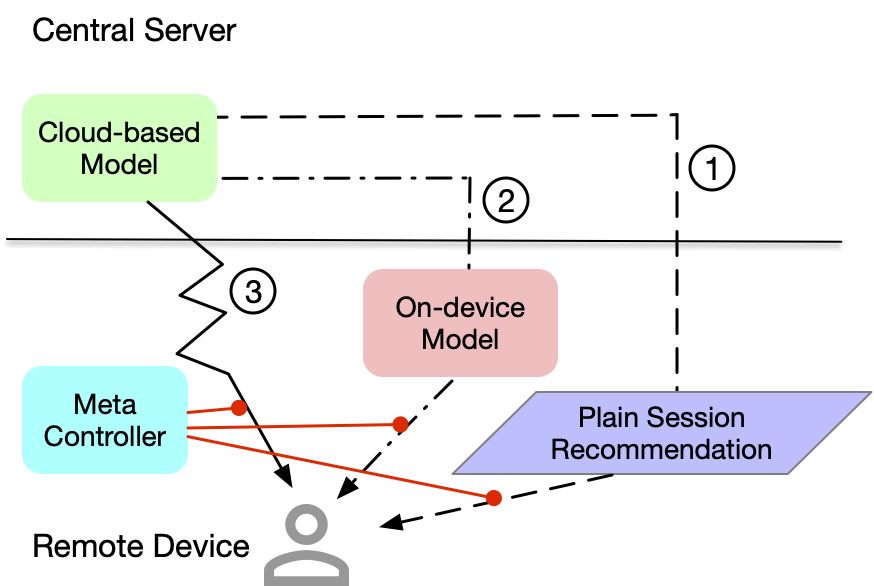}
  \caption{\textcircled{1} The conventional session recommendation, where the cloud-based recommender is unaware of the real-time user interactions in the session. \textcircled{2} The previous design that leverages the on-device model to refine the recommendation results from the cloud model based on the in-session behaviors. \textcircled{3} A real-time but expensive in-session recommendation re-fresh from the cloud-based model. Our work is to construct a meta controller that dynamically manages the collaboration among three recommendation mechanisms.}\label{fig:motivation}
\end{figure}

The popular industrial recommender systems mainly benefit from the rapid development of cloud computing~\cite{duc2019machine}, which constructs a cost-efficient computing paradigm to serve the billion-scale customers. The inherent algorithms progressively evolved from collaborative filtering~\cite{sarwar2001item,zhao2010user}, matrix factorization~\cite{koren2009matrix} to deep learning based methods~\cite{cheng2016wide,guo2017deepfm, zhou2018deep,sun2019bert4rec,tan2021sparse} \textit{etc.}, which explore the gain of the low-level feature interaction and high-level semantic understanding. Although the performance is improved by more powerful models, the corresponding computational cost increases concomitantly, requiring the mechanism adjustment of recommendation~\cite{wang2021survey} and more expensive accelerations in hardware~\cite{norrie2021design}. An exemplar is the E-commerce session recommendation~\cite{jannach2017session} that ranks the top-K candidates in one session to improve the efficiency and reduce the computational burden in the cloud server, especially for the infinite feeds recommendation. The path \textcircled{1} in Figure~\ref{fig:motivation} illustrates this type of recommendation in the industrial scenarios, and when the length of session is 1 in the extreme, \textit{i.e.,} one item in once recommendation, it becomes path \textcircled{3} in Figure~\ref{fig:motivation}, a prohibitively high re-fresh. Note that, the mechanism of the path \textcircled{3} is usually better than that of the path \textcircled{1}, since it senses the user interaction feedback more frequently.  However, it is not scalable due to the bottleneck of the computational resource towards billions of customers, and thus the re-fresh only slightly applies in practice when it is really indispensable compared to other surrogate ways. 

The recent development in edge computing~\cite{zhou2019edge,yao2021edge} drives the AI models deployed in remote devices to reduce the communication cost and incorporate the fine-grained real-time features. Specifically, for recommendation, there are several explorations~\cite{yang2020federated,sun2020generic,gong2020edgerec,yao2021dccl} to leverage the remote computing power for recommender systems, with the increasing capacity of smart phones and the aid of softwares including TFLite\footnote{\url{https://www.tensorflow.org/lite}}, PyTorch Mobile\footnote{\url{https://pytorch.org/mobile/home/}} and  CoreML\footnote{\url{https://developer.apple.com/documentation/coreml}}. Different from the rich resources of the cloud server in the aspects of storage, energy and computing speed, on-device recommenders must be lightweight by applying the compression techniques~\cite{sun2020generic,cai2020tinytl} or the split-deployment methods~\cite{gong2020edgerec,banitalebi2021auto}. However, 
compared to the above constraints, the advantages of on-device recommenders in real-time and personalization are also significant to the E-commerce session recommendation, \textit{e.g.,} EdgeRec~\cite{gong2020edgerec} and DCCL~\cite{yao2021dccl}. Beyond the developed algorithms, the common mechanism behind all aforementioned on-device recommenders is sequentially combined with the cloud-based recommendation, which is illustrated in the path \textcircled{2} of Figure~\ref{fig:motivation}. Namely, the on-device model re-ranks the unexposed candidates recommended by the cloud-based model with real-time user interactive features. However, one shortcoming of this mechanism is the number of cached items due to the limit of mobile storage that cannot satisfy dramatic user interest changes.

From the above discussion, we know the current cloud-based and on-device recommenders both have their pros and cons. Yet according to their characteristics, they actually can be complementary to construct a more flexible mechanism for recommendation: When the user interests almost have no changes, the plain session recommendation is enough without invoking on-device recommender to frequently refine the ranking list; When the user interests slightly change, re-ranking the candidates from the item cache by the on-device model according to the real-time features can improve the accuracy of recommendation; When the user interests dramatically change, it will be better to explore more possible candidates from the large-scale item pool by the cloud-based re-fresh. This introduces the idea of ``Meta Controller" to achieve the \emph{Device-Cloud Collaboration} shown in Figure~\ref{fig:motivation} (a similar and concurrent two-mode complementary design can refer to AdaRequest~\cite{qian2022intelligent}). Despite straightforward, it is challenging to collect the training samples for the meta controller, since three mechanisms in Figure~\ref{fig:motivation} are separately executed and so are the observational samples. Naively deploying a random-guess model as a surrogate online to collect the data sometimes is risky or requires a lengthy exploration. To solve this dilemma, we introduce a novel idea from causal inference to compare the causal effect~\cite{kunzel2019metalearners,johansson2016learning,shalit2017estimating} of three recommendation mechanisms, which implements an efficient offline sample construction and facilitates the training of meta controller. In a nutshell, the contribution of this paper can be summarized as follows.
\begin{itemize}
    \item Considering the pros and cons of the current cloud-based recommender and the on-device recommender, we are among the first attempts to propose a meta controller to explore the ``Device-Cloud Collaborative Recommendation".
    \item To overcome the dataset absence issue when training the meta controller, we propose a dual-uplift modeling method to compare the causal effect of three recommendation mechanisms for an efficient offline sample construction.
    \item Regarding the constrained resources for the cloud-based recommendation re-fresh, we propose a label smoothing method to suppresses the prediction of meta controller during the training to adapt the practical requirements.
\end{itemize}
The extensive experiments on a range of datasets demonstrate the promise of the meta controller in Device-Cloud Collaboration.

\section{Related Works}
\subsection{Cloud-Based Recommendation}
Recommendation on the basis of cloud computing~\cite{duc2019machine} has achieved great success in the industrial companies like
Microsoft, Amazon, Netflix, Youtube and Alibaba \textit{etc}.
A range of  algorithms from the low-level feature interaction to the high-level semantic understanding have been developed in the last decades~\cite{yao2017joint,yao2017discovering}. Some popular methods like Wide$\&$Deep~\cite{cheng2016wide} and DeepFM~\cite{guo2017deepfm} explore the combination among features to construct the high-order information, which reduces the burden of the subsequent classifier. Several approaches like DeepFM~\cite{xue2017deep}, DIN~\cite{zhou2018deep} and Bert4Rec~\cite{sun2019bert4rec} resort to deep learning to hierarchically extract the semantics about user interests for recommendation~\cite{pan2021click,zhang2022re4}. More recent works like MIMN~\cite{pi2019practice}, SIM~\cite{pi2020search} and Linformer~\cite{wang2020linformer} solve the scalability and efficiency issues in comprehensively understanding user intention from lifelong sequences. Although the impressive performance has been achieved in various industrial recommendation applications, almost all of them depends on the rich computing power of the cloud server to support their computational cost~\cite{sun2020convergence}.

\subsection{On-Device Recommendation}
The cloud-based paradigm greatly depends on the network conditions and the transmission delay to respond to users' interaction in real time~\cite{yao2021edge}. With the increasing capacity of smart phones, it becomes possible to alleviate this problem by deploying some lightweight AI models on the remote side to save the communication cost and access some real-time features~\cite{zhou2019edge}. Previous works in the area of computer vision and natural language processing have explored the possible network compression techniques, \textit{e.g.,} MobileNet~\cite{sandler2018mobilenetv2} and MobileBert~\cite{sun2020mobilebert}, and some neural architecture search approaches~\cite{zoph2016neural,cai2018efficient}. The recent CpRec for on-device recommendation studied the corresponding compression by leveraging the adaptive decomposition and the parameter sharing scheme~\cite{sun2020generic}. In comparison, EdgeRec adopted a split-deployment strategy to reduce the model size without compression, which distributed the huge id embeddings on the cloud server and sent them to the device side along with items~\cite{gong2020edgerec}. Some other works, DCCL~\cite{yao2021dccl} and FedRec~\cite{yang2020federated}, respectively leveraged the on-device learning for the model personalization and the data privacy protection. Nevertheless, the design of the on-device recommendation like EdgeRec and DCCL is sequentially appended after the cloud-based recommendation results, which might be stuck by the limited item cache.

\subsection{Causal Inference}
Causal inference has been an important research topic complementary to the current machine learning area, which improves the generalization ability of AI models based on the limited data or from one problem to the next~\cite{scholkopf2021toward}. The broadly review of the causality can be referred to this literature~\cite{pearl2018book} and in this paper, we specifically focus on the counterfactual inference to estimate the heterogeneous treatment effect~\cite{kunzel2019metalearners}. Early studies mainly start from the statistical perspective like S-Learner and X-Learner~\cite{kunzel2019metalearners}, which alleviates the treatment imbalance issue and leverages the structural knowledge. Recent works explore the combination with deep learning to leverage the advantages of DNNs in representation learning~\cite{johansson2016learning,shalit2017estimating,scholkopf2021toward}. In the industrial recommendation, the idea of causal inference is widely applied for the online A/B testing, debiasing~\cite{zhang2021causerec} and generalization~\cite{kuang2020causal}. Except above works, the most related methods about our model are uplift modeling, which is extensively used to estimate the causal effect of each treatment~\cite{olaya2020survey}. Specifically, the constrained uplift modeling is practically important to meet the requirements of the restricted constraint for each treatment~\cite{zhao2019uplift}. In this paper, we are first to explore its application in solving the dataset absence issue of training the meta controller for device-cloud collaborative recommendation.

\section{The Proposed Approach}
Given a fixed-length set of historical click items $\bfH_{\cloud,i}$ that the $i$-th user interacts before the current session, and all side features (\textit{e.g.,} statistical features) $\bfS_{\cloud,i}$, the cloud-based recommendation is to build the following prediction model that evaluates the probability for the current user interests regarding the $j$-th candidate $\bfX_{j,i}$,
$$f_\cloud\left(\bfX_{j,i}\big|\bfH_{\cloud,i}, \bfS_{\cloud,i}\right).$$
For the on-device recommendation, the model could leverage the real-time interactions and other fine-grained features (\textit{e.g.,} exposure and prior prediction score from the cloud-based recommendation) in the current session, denoting as $\bfH_{\device,i}=\bfH_{\cloud,i}\cup \bfH_{\text{add},i}$ and $\bfS_{\device,i}$ respectively. Similarly, we formulate the on-device prediction model regarding the $j$-th candidate $\bfX_{j,i}$ as follows,
$$f_\device\left(\bfX_{j,i}\big|\bfH_{\device,i}, \bfS_{\device,i}\right).$$
As for the recommendation re-fresh, considering its consistency with the common session recommendation, the 
model actually is shared with the cloud-based recommendation but accessing the richer $\bfH_{\device,i}$ instead of $\bfH_{\cloud,i}$ as input features when invoking. That is, the prediction from the re-fresh mechanism is as follows,
$$f_\cloud\left(\bfX_{j,i}\big|\bfH_{\device,i}, \bfS_{\cloud,i}\right).$$
The gain of this mechanism is from the feature perspective instead of the model perspective, namely, the real-time feedback about user interactions in the current session $\bfH_{\device,i}$. Specially, we should note that in practice, it directly use the same model $f_\cloud(\cdot)$ without additionally training a new model except the input difference. 

\subsection{Meta Controller}
Given the aforementioned definitions, our goal in this paper is to construct a meta controller that manages the collaboration of the above mechanisms to more accurately recommend the items according to the user interests. Let $f_\meta(\cdot)$ represents the model of meta controller, which inputs the same features as $f_\device(\cdot)$ and is deployed in the device. Assuming $t=0,~1,~2$ represents invoking three recommendation mechanisms in Figure~\ref{fig:motivation} respectively, we expect $f_\meta(\cdot)$ makes a decision to minimize the following problem.
\begin{align} \label{eq:meta}
    \min_{f_\meta} \mathbb{E}_{(\bfX,\bfY)\sim P_\meta(\bfX,\bfY)}\left[\ell\left(\bfY, \sum_{t=0}^2 f_\meta(\bfX)f_t\left(\bfX\right)\right)\right],
\end{align}
where $P_\meta(\bfX,\bfY)$ is the underlying distribution about the user interaction $\bfY$ (click or not) on the item $\bfX$, $\ell$ is the cross-entropy loss function about the label $\bfY$ and the model prediction, and  $f_t(\cdot)$ corresponds to one of previous recommendation mechanisms based on the choice $t$. Although the above Eq.~\eqref{eq:meta} resembles learning a Mixture-of-Expert problem~\cite{shazeer2017outrageously}, it is different as we only learn $f_\meta(\cdot)$ excluding all pre-trained $f_t(\cdot)$. In addition, this problem is specially challenging, since we do not have the samples from the unknown distribution $P_\meta(\bfX,\bfY)$ to train the meta controller. 

\paragraph{\textbf{Dataset Absence Issue}} The generally collected training samples in the industrial recommendation are from the scenarios where the stable recommendation model has been deployed. For example, we can acquire the training samples $\{(x,y)\}_\cloud$ under the serving of an early-trained session recommendation model, or $\{(x,y)\}_\device$ under the serving of the on-device recommender. However, in the initialing point, we do not have a well-optimized meta controller model that can deploy for online serving and collect the training samples under the distribution $P_\meta(\bfX,\bfY)$. Naively deploying a random-guess model or learning from the scratch online sometimes can be risky or require a lengthy exploration to reach a stationary stage. In the following, we propose a novel offline dataset construction method from the perspective of causal inference~\cite{kunzel2019metalearners,johansson2016learning,shalit2017estimating}.

\begin{figure*}[!t]
  \centering
  \includegraphics[width=0.82\linewidth]{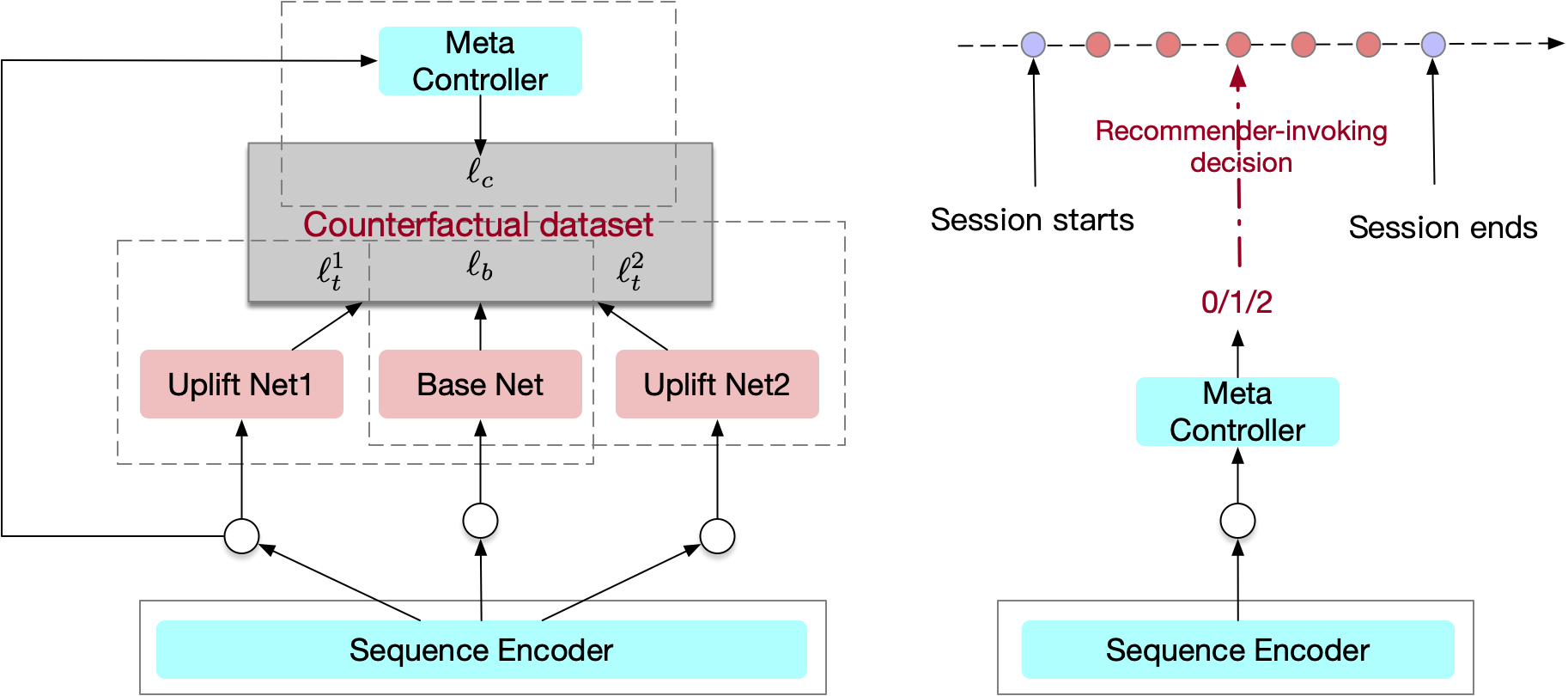}
  \caption{The dual-uplift modeling method (the left panel) that compares the causal effect of recommendation mechanisms for the efficient offline sample construction, and then we can train the meta controller on the generated counterfactual dataset. When serving in the session, meta controller makes a decision to invoke the corresponding recommender (the right panel).}\label{fig:model}
\end{figure*}

\subsection{Causal Inference for Sample Construction}
One important branch of causal inference is counterfactual inference under interventions~\cite{pearl2018book}, which evaluates the causal effect of one treatment based on the previous observational studies under different treatments. This coincidentally corresponds to the case in device-cloud collaboration to train the meta controller, where we can collect the training samples $\mathcal{D}_t$ of three different scenarios, the cloud-based session recommendation, the on-device recommendation, the cloud-based re-fresh, denoting as follows,
$$
    \mathcal{D}_t = \begin{cases}
    \{(\bfH_{\cloud},\bfO(t))\}_\cloud,  & t=0 \\
    \{(\bfH_{\device},\bfO(t))\}_\device,  & t=1 \\
    \{(\bfH_{\device},\bfO(t))\}_\text{re-fresh}, & t=2 ~.
    \end{cases} \\
$$
Note that, $\bfH_{(.)}$ is the user interaction in three scenarios (same to previous definitions), and $\bfO(\cdot)$ denotes the binary outcome of $\bfH_{(.)}$, which means under the treatment $t$ whether the user will click in the following $L$ exposed items $\{(x_{l},y_{l})\}_{l \in (1,2,...,L)}$ in the same session. With these three heterogeneous sample sets, we can leverage the idea of counterfactual inference to compare the causal effect of recommendation mechanisms and construct the training samples of meta controller. To be simple, consider the following estimation of Conditional Average Treatment Effect (CATE)$$
\textbf{CATE:} \quad \begin{cases} 
\tau_\cloud(\bfH) = 0 ~(\text{session recommendation as the base})\\
\tau_\device(\bfH) = \mathbb{E}\left[\bfO(1)|\bfH\right] - \mathbb{E}\left[\bfO(0)|\bfH\right] \\
\tau_\refresh(\bfH) = \mathbb{E}\left[\bfO(2)|\bfH\right] - \mathbb{E}\left[\bfO(0)|\bfH\right] \\
\end{cases}
$$
where we can directly treat the regular cloud-based session recommendation as the control group, on-device recommendation and re-fresh as the treatment groups. Then, by separately training a supervised model for each group, we can approximately compute the CATE of two treatments via T-learner or X-learner~\cite{kunzel2019metalearners}. Specially, the recent progress in the uplift modeling combined with deep learning~\cite{johansson2016learning,shalit2017estimating} shows the promise about the gain in the homogeneous representation learning. Empirically, we can share a backbone of representation learning and 
the base output network among different supervised models~\cite{wenwei2021}. 

In Figure~\ref{fig:model}, we illustrate the proposed dual-uplift modeling method that constructs the counterfactual samples to train the meta controller. As shown in the figure, we build two uplift networks along with the base network to respectively estimate the CATE of two treatments, and the base network is to build the baseline for the data $\mathcal{D}_0$ in the control group. Formally, the base network is defined in the following to align our illustration in Figure~\ref{fig:model},
$$
    g_\cloud(\cdot) = g_{\base}(\cdot)\circ \phi_{\seq}(\cdot),
$$
where $\phi_\seq(\cdot)$ is the encoder for the historical click interactions $\bfH_\cloud$ or $\bfH_\device$ and $\circ$ is the function composition operator\footnote{\url{https://en.wikipedia.org/wiki/Function_composition}}, and $g_\base$ is named as the outcome network of the baseline session recommendation in the taxonomy of causal inference~\cite{shalit2017estimating}. Similarly, for the treatment effect of the on-device recommendation, we follow the design of $f_\device$ to construct an uplift network $g_\uplifta(\cdot)$ to fit the data $\mathcal{D}_1$. As shown in Figure~\ref{fig:model}, its formulation is as follows,
$$
    g_\device(\cdot) = g_{\uplifta}(\cdot)\circ \phi_{\seq}(\cdot),
$$
where $g_\uplifta(\cdot)$ is the outcome network of the on-device recommendation. Note that, $g_\cloud(\cdot)$ and $g_\device(\cdot)$ share the same feature encoder $\phi_\seq(\cdot)$ to make the representation more homogeneous that satisfies the assumption of the causal inference. Finally, for the re-fresh mechanism, we use another uplift network to fit its data $\mathcal{D}_2$, which is denoted by the following equation,
$$
    g_\refresh(\cdot) = g_{\upliftb}(\cdot)\circ \phi_{\seq}(\cdot).
$$
Note that, although the features of $g_\cloud(\cdot)$ and $g_\refresh(\cdot)$ have the same forms, they are actually non-independent-identically-distributed (non-IID) due to different invoking frequencies. Therefore, we use $g_\upliftb(\cdot)$ instead of  $g_\cloud(\cdot)$ to more flexibly model the outcome of the cloud-based re-fresh mechanism. With the above definitions, we can solve the following optimization problems to learn all surrogate models $g_\cloud(\cdot)$, $g_\device(\cdot)$ and $g_\refresh(\cdot)$.
\begin{align} \label{eq:surrogate_learning}
    \begin{cases}
    \min \frac{1}{|\mathcal{D}_0|}\sum_{(h,o)\in \mathcal{D}_0}\ell_b(o,~g_\cloud(h)) \\
    \min \frac{1}{|\mathcal{D}_1|}\sum_{(h,o)\in \mathcal{D}_1}\ell_t^1(o,~g_\device(h)) \\
    \min \frac{1}{|\mathcal{D}_2|}\sum_{(h,o)\in \mathcal{D}_2}\ell_t^2(o,~g_\refresh(h)),
    \end{cases}
\end{align}
where $\ell_b(\cdot)$, $\ell_t^1(\cdot)$ and $\ell_t^2(\cdot)$ are the cross-entropy loss. After learning $g_\cloud(\cdot)$, $g_\device(\cdot)$ and $g_\refresh(\cdot)$, we can acquire a new counterfactual training dataset by comparing the CATE of each sample under three mechanisms defined as follows.
\begin{align} \label{eq:sample_construction}
\begin{split}
    \mathcal{D}_\meta = \left\{(h, c)\big| \right. & c=\text{IND}(\tau_\cloud(h), \tau_\device(h), \tau_\refresh(h)), \\ 
    & \qquad \left.\forall (h,o)\in\mathcal{D}_0\cup\mathcal{D}_1\cup\mathcal{D}_2\right\}
\end{split}
\end{align}
where $\text{IND}(\cdot,\cdot,\cdot)$ means the position indicator ($[1,0,0]^\top$, $[0,1,0]^\top$, or $[0,0,1]^\top$) of the maximum in three CATEs, $\tau_\device(h)=g_\device(h)-g_\cloud(h)$ and $\tau_\refresh(h)=g_\refresh(h)-g_\cloud(h)$ if T-Learner is used. Here, the counterfactual inference is used to compute the CATE of each sample collected under different treatments. Note that, if X-Learner is used, CATE can be computed in a more comprehensive but complex manner and we kindly refer to this work~\cite{kunzel2019metalearners}. 

\textbf{Remark about the difference of $f(\cdot)$ and $g(\cdot)$}. Previous paragraphs mentioned different parameterized models \textit{e.g.,} $f_\cloud(\cdot)$ and $g_\cloud(\cdot)$, which could be classified into two categories by $f(\cdot)$ and $g(\cdot)$. Their biggest difference is whether these models will be used online. For the models described in Preliminary, they are all real-world models serving online and in most cases are developed and maintained by different teams in the industrial scenarios. For the $g(\cdot)$-style models described in this paragraph, they are all surrogate models, which are only used for the sample construction and not deployed online. Roughly, it will be better if they can be consistent with the well-optimized $f(\cdot)$-style models, but in some extreme cases they actually can be very different only if the approximate CATE is accurate enough to make the constructed $\mathcal{D}_\meta$ contain the sufficient statistics about the distribution $P_\meta$ in Eq.~\eqref{eq:meta}.

\subsection{Training the Meta Controller}
Generally, it is straightforward to train the meta controller by a single-label multi-class empirical risk minimization with the aid of the counterfactual dataset $\mathcal{D}_\meta$ in Eq.~\eqref{eq:sample_construction} as follows,
\begin{align}\label{eq:erm}
    \min_{f_\meta} \frac{1}{|\mathcal{D}_\meta|} \sum_{(h,c)\in \mathcal{D}_\meta}\ell\left(c,~f_\meta(h)\right).
\end{align}
Unfortunately, not all recommendation mechanisms in Figure~\ref{fig:motivation} are cost-effective. Specially, the cloud-based recommendation re-fresh usually suffers from the restricted computing resources in the infinite feeds recommendation, yielding a hard invoking constraint in practice. This then introduces a constrained optimization problem in device-cloud collaboration and breaks the regular training of meta controller like Eq.~\eqref{eq:erm}. Actually, this is a classical constrained budget allocation problem, which has been well studied in last decades~\cite{zhao2019unified,meister2020generalized}. In this paper, we introduces a Label Smoothing (LS) solution to generally train the meta controller under the specific constraint on the cloud-based recommendation re-fresh. In detail, given the maximal invoking ratio $\epsilon$ about the cloud-based re-fresh, we make the relaxation for Eq.~\eqref{eq:erm} via the label smoothing.
\begin{align}\label{eq:rerm}
\begin{split}
    & \min_{f_\meta} \frac{1}{|\mathcal{D}_\meta|} \sum_{(h,c)\in \mathcal{D}_\meta}\ell\left(\hat{c},~f_\meta(h)\right), \\
    & \text{where}~\hat{c}=\begin{cases}
    (1-\lambda)c + \lambda c^-, & c=[0,0,1]^\top~\&~\text{over budget}  \epsilon\\
    c, & \text{Default} \\
    \end{cases}
\end{split}
\end{align}
where we use $\hat{c}$ to substitute $c$ to supervise the model training and $\hat{c}$ is smoothed when the ratio of the batch of predictions is beyond $\epsilon$, $c^-$ is the one-hot vector corresponding to the second largest prediction in $f_\meta(h)$. The relaxation in Eq.~\eqref{eq:rerm} means when the invoking ratio is over-reached, the supervision should be biased to the second possible choice $c^-$ for the prediction of $c=[0,0,1]^\top$. Note that, when the hard ratio $\epsilon$ is not reached, $\hat{c}$ still keeps the original label $c$ in the default case, namely degenerates to Eq.~\eqref{eq:erm}. Note that, Eq.~\eqref{eq:rerm} can be easily implemented in the stochastic training manner and after training until the invoking ratio is almost under $\epsilon$, we acquire the final meta controller that can be deployed online in the device-cloud collaborative recommendation. Up to now, we describe all the contents of our meta controller method and for clarity, we summarize the whole framework in Algorithm~\ref{algo:meta}.

\begin{algorithm}[!t]
\SetAlgoLined
The datasets $\mathcal{D}_0$, $\mathcal{D}_1$ and $\mathcal{D}_2$ under three mechanisms, the meta controller $f_\meta(\cdot)$ and the re-fresh constraint $\epsilon$.\\
\textbf{Phase I:}~\colorbox{gray!30}{$\rhd$~\emph{Dataset Construction}} \\
\qquad 1) Initialize models $g_\cloud(\cdot)$, $g_\device(\cdot)$ and $g_\refresh(\cdot)$. \\
\qquad 2) Learn these three surrogate models via Eq.~\eqref{eq:surrogate_learning}. \\
\qquad 3) Compute CATEs of each sample in $\mathcal{D}_0$, $\mathcal{D}_1$ and $\mathcal{D}_2$. \\
\qquad 4) Compare all CATEs via Eq.~\eqref{eq:sample_construction} to get $\mathcal{D}_\meta$. \\
Output the dataset $\mathcal{D}_\meta$.\\
\textbf{Phase II:}~\colorbox{gray!30}{$\rhd$~\emph{Training Stage}} \\
\While{not converging}{
\quad 1) Hook a mini-batch of samples from $\mathcal{D}_\meta$. \\
\quad 2) Forward samples to $f_\meta(\cdot)$ to compute $\hat{c}$ in Eq.\eqref{eq:rerm}. \\ 
\quad 3) Backward the loss in Eq.~\eqref{eq:rerm} to update $f_\meta(\cdot)$.
}
Output the Meta Controller $f_\meta(\cdot)$.
\caption{\mbox{Meta Controller}}
\label{algo:meta}
\end{algorithm}

\section{Experiments}
\begin{figure*}[!ht]
  \centering
  \includegraphics[width=0.24\linewidth]{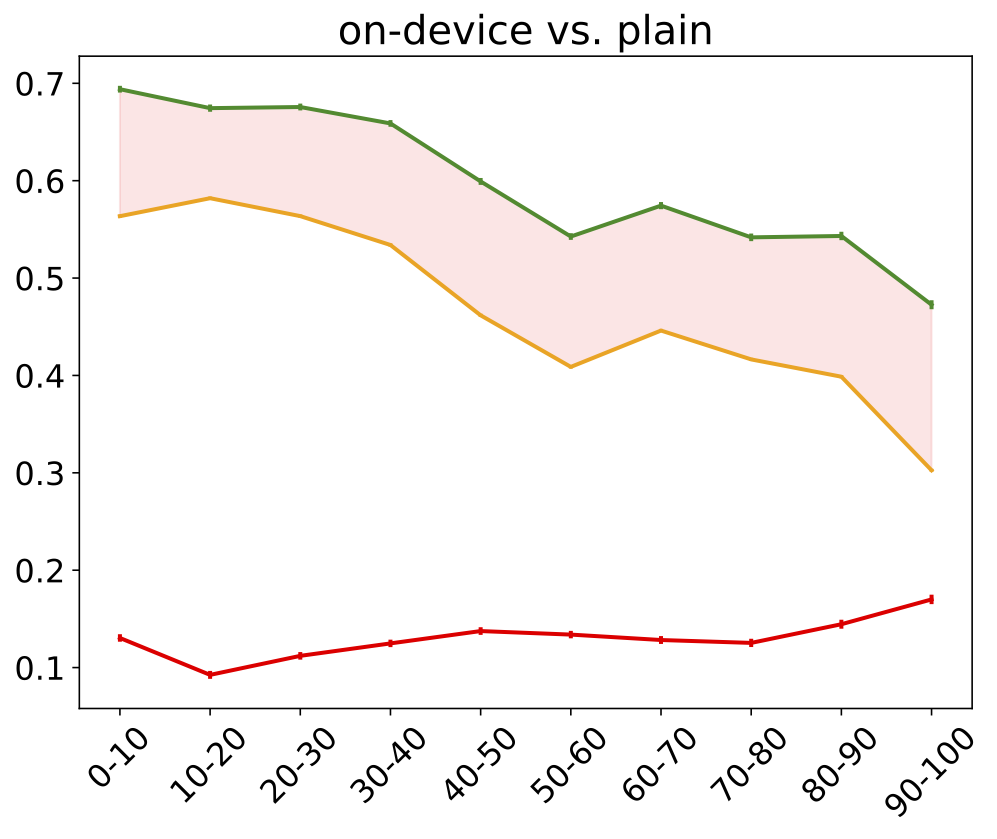}
  \hspace{0.5cm}
  \includegraphics[width=0.24\linewidth]{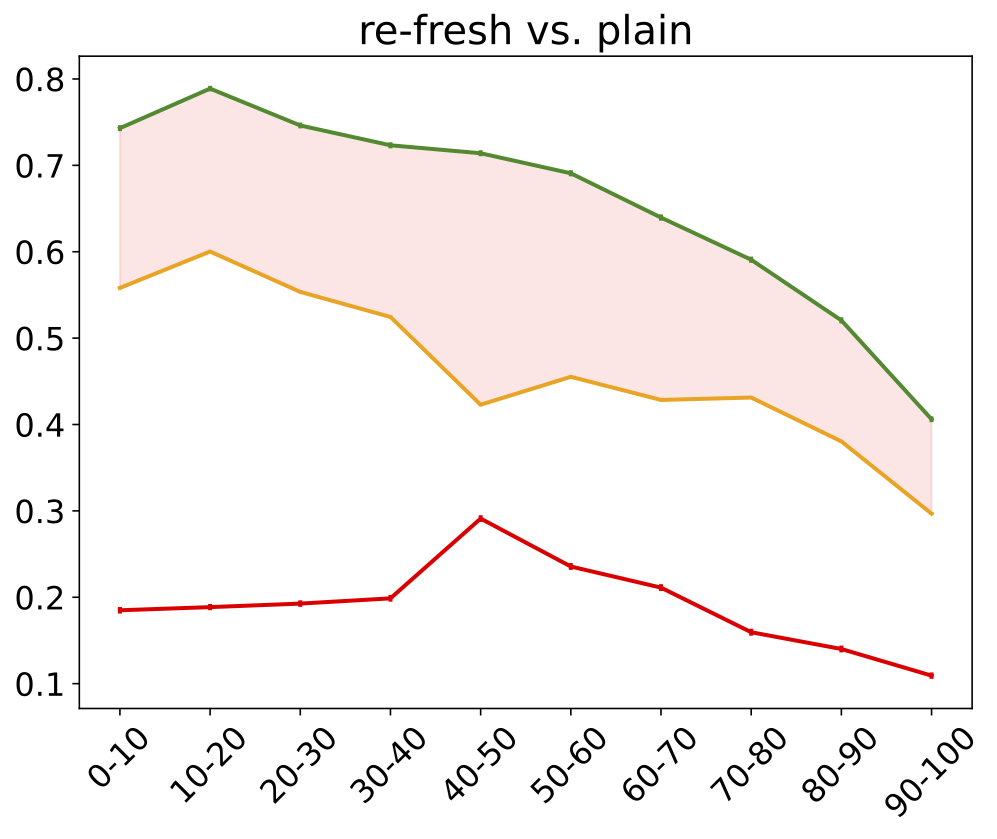}
  \hspace{0.5cm}
  \includegraphics[width=0.24\linewidth]{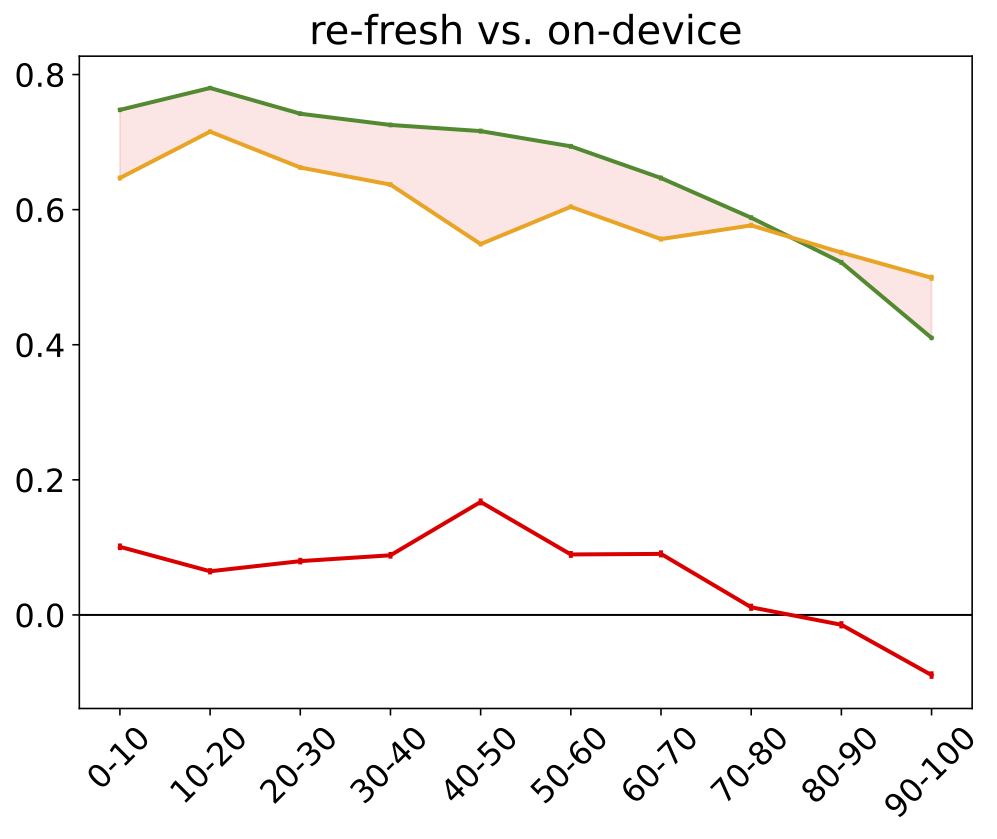}
  \caption{The uplift score (y-axis) at each percentile (x-axis) on the Alipay dataset. The red line is the uplift gain, which is computed by the difference between the treatment response rate (the green line) and the control response rate (the yellow line) at each percentile. It reflects how one treatment performs compared to its baseline in different granular groups.}\label{fig:auuc}
\end{figure*}
In this section, we conduct a range of experiments to demonstrate the effectiveness of the proposed framework. To be specific, we will answer the following questions in our device-cloud collaboration.
\begin{enumerate}
    \item \textbf{RQ1:} Whether the proposed method can help train a meta controller to achieve a better performance via device-cloud collaboration compared to previous three types of mechanism? To our best knowledge, we are the first attempt to explore this type of collaboration between cloud-based recommenders and the on-device recommender.
    \item \textbf{RQ2:} Whether the proposed method can adapt to the practical resource constraint after the training of the meta controller by means of Eq.~\eqref{eq:rerm}. It is important to understand the underlying trade-off in performance under the practical constraint in the device-cloud collaboration.
    \item \textbf{RQ3:} How is the potential degeneration from the collaboration of three recommendation mechanisms to the arbitrary two of them, and how is the ablation study that comprehensively analyzes the exemplar intuition support, the training procedure, the inherit parameter characteristics?
\end{enumerate}

\subsection{Experimental Setup}
\subsubsection{Datasets} We implement the experiments on two public datasets Movienlens and Amazon, and one real-world large-scale backgroud word dataset, termed as Alipay for simplicity. The statistics of three datasets are summarized in Table~\ref{table:Statistics}. Generally, each dataset contains five training sets, \textit{i.e.,} two sets to train the base cloud recommender and on-device recommender respectively, three sets to learn the meta controller. As for the test sets of each dataset, one set is used to measure the CATE of meta controller, and the other one is to measure the click-through rate (CTR) performance of the models. Note that, Movielens and Amazon do not have the session information, and we thus manually simulate the session to meet the setting of this study. More details about the simulations of Movielens and Amazon are kindly referred to the appendix.

\begin{table}[!t]
\caption{Statistics of datasets used in the training. On the Alipay dataset, M denotes million and K denotes thousand.}
\small
\centering
\begin{tabular}{l|c|c|c|c}
\toprule
    Dataset & Type & \#user & \#item  & \#sample \\ 
\midrule
\multirow{5}{*}{Movielens} & CTR-Cloud & & &7,900,245 \\
& CTR-Device & & & 26,998,425 \\
& CATE-Cloud & 6,040 & 3,706 & 4,819,845\\
& CATE-Device & & & 4,819,845\\
& CATE-Refresh & & & 4,819,845\\
\midrule
\multirow{5}{*}{Amazon} & CTR-Cloud & & &17,444,825 \\
& CTR-Device & & &285,809,710 \\
& CATE-Cloud &49,814 & 36,997&2,588,895 \\
& CATE-Device & & & 2,588,895\\
& CATE-Refresh & & & 2,588,895\\
\midrule
\multirow{5}{*}{Alipay} & CTR-Cloud & 10.5M & \multirow{5}{*}{10K} & 31M \\
& CTR-Device & 5.4M &  & 14M \\
& CATE-Cloud & 0.6M & & 1M \\
& CATE-Device & 66K & & 0.2M \\
& CATE-Refresh & 0.3M & & 0.3M \\
\bottomrule
\end{tabular} \label{table:Statistics}
\end{table}

\subsubsection{Baselines \& Our methods}
We compare the meta controller mechanism with previous three existing mechanisms, \textit{i.e.,} the cloud-based session recommendation, the on-device recommendation and the cloud-based re-fresh. In three mechanisms, we all leverage DIN~\cite{zhou2018deep} as the backbone, which takes the candidates as the query \textit{w.r.t.} the user
historical behaviors to learn a user representation for prediction, but it correspondingly incorporates different features as in $f_\cloud$, $f_\device$ and $f_\refresh$. To be simple, we term the models in  previous three mechanisms respectively  as $\bf{CRec}$ (cloud-based session recommender), $\bf{CRRec}$ (cloud-based re-fresh) and $\bf{ORec}$ (on-device recommender), and we call our method $\bf{MCRec}$ (the Meta-Controller-based recommender).  Besides, we also provide a baseline that randomly mixes three types of the existing recommendation mechanisms, termed as $\bf{RMixRec}$. For the network structure details \textit{e.g.,} the dimensions and the training schedules of DINs in CRec, CRRec, ORec, RMixRec and MCRec, please refer to the appendix.

\begin{table*}[!t]
	\caption{Recommendation performance of previous recommendation mechanisms and our method under different $\epsilon$ constraints.}
	\label{tab:ctr}
	\centering
	{
	\begin{tabular}{l | c c c | c c c | c c c}
	\toprule
	\multirow{2}{*}{Methods} & \multicolumn{3}{c|}{Movielens} & \multicolumn{3}{c|}{Amazon} & \multicolumn{3}{c}{Alipay} \\
	& HitRate$@$1 & NDCG$@$5 & AUC &  HitRate$@$1 & NDCG$@$5 & AUC & HitRate$@$1 & NDCG$@$5 & AUC \\
	\midrule
	CRec &   32.22&50.13 &92.46 & 16.27 & 27.36 &77.04  & 17.85 & 34.96 & 72.39 \\
	ORec  &  36.98&55.31 &93.70 &17.31 & 28.94 & 77.95 & 17.88 & 35.02 & 72.80 \\
	CRRec  & 37.28 & 55.85 &93.88 &17.56 & 29.24 &78.98& 18.09 & 35.41 & 73.35 \\
	RMixRec  & 35.52 & 53.76 &93.34 & 17.06 & 28.51 & 77.99 & 17.90 & 35.06 & 72.83 \\
	\midrule
	MCRec (w/o constraint)   & {37.84} & \textbf{56.20} &\textbf{93.90}  & \textbf{17.70} & \textbf{29.48} &\textbf{79.05} & \bf{18.26} & \bf{35.77} & \bf{73.86} \\
	MCRec ($\epsilon=0.5$)  & \textbf{37.87} & 56.19 & 93.88  &17.68 & 29.48 & 79.03 & 18.14 & 35.62 & 73.45 \\
	MCRec ($\epsilon=0.3$)  & 37.62 & 55.88 & 93.80  &17.61  & 29.38 & 78.79 & 17.98 & 35.33 & 73.32 \\
	MCRec ($\epsilon=0.1$)  & 37.16& 55.45 & 93.67  & 17.50 & 29.21 & 78.38 & 17.91 & 35.17 & 73.03 \\
	\bottomrule
	\end{tabular}
	}
\end{table*}

\subsubsection{Evaluation Protocols}
The process of training Meta Controller consists of two phases as summarized in Algorithm~\ref{algo:meta}, and thus the metrics we used in this paper will evaluate these two phases respectively. For the phase of the dataset construction, we mainly measure the causal effect, namely CATE of the synthetic dataset based on the counterfactual inference. While for the training phase, we mainly evaluate the click-through rate (CTR) performance of the recommenders including baselines and our method. For the dataset construction stage, we use the AUUC~\cite{diemert2018large} and QINI coefficients~\cite{Radcliffe2007UsingCG} as the evaluation metrics, which are defined as follows,
\begin{align} 
\begin{split} 
    & \text{AUUC} = \frac{1}{2N}\sum_{n=1}^N \left[ p(n)\text{Diff}(n)-p(n-1)\text{Diff}(n-1)\right]\\
    & \text{QINI} = \text{AUUC} - \text{AUUC}_\text{random},\nonumber  
\end{split}
\end{align}
where $\text{Diff}(n)$ calculates the relative gain from the top-n control group to the top-n treatment group (sorted by prediction scores)~\cite{diemert2018large}, $p(n)=n/N$ is the normalizing ratio, $\text{AUUC}_\text{random}$ is a random sort baseline to characterize the data randomness.
Regarding the final recommendation performance, first, we give some notations that $\mathcal{U}$ is the user set, $\mathds{1}(\cdot)$ is the indicator function, $R_{u,g_u}$ is the rank generated by the model for the ground truth item $g_u$ and user $u$, $f$ is the model to be evaluated and $D_T^{(u)}$, $D_F^{(u)}$ is the positive and negative sample sets in testing data. Then, the widely used HitRate, AUC
and NDCG respectively is used, which are defined by the following equations.
\begin{align} 
\begin{split} 
    & \text{HitRate}@K = \frac{1}{|\mathcal{U}|}\sum_{u\in \mathcal{U}} \mathds{1}(R_{u,g_u}\leq K),\\
    & \text{NDCG}@K = \sum_{u\in \mathcal{U}}  \frac{1}{|\mathcal{U}|}  \frac{2^{\mathds{1}(R_{u,g_u}\leq K)}-1}{\log_2(\mathds{1}(R_{u,g_u}\leq K)+1)},\\
    & \text{AUC} = \sum_{u\in\mathcal{U}} \frac{1}{|\mathcal{U}|} \frac{\sum_{x_0\in D_T^{(u)}} \sum_{x_1 \in D_F^{(u)}}\mathbf{1}[f(x_1)<f(x_0)]}{|D_T^{(u)}||D_F^{(u)}|} , \nonumber
\end{split}
\end{align}

\subsection{Experimental Results and Analysis}

\begin{table}[!ht]
    \centering
    \caption{The AUUC scores and the QINI coefficients of different recommendation mechanisms. Note that, CRec performs as baseline in the control group relative to ORec and CRRec in the treatment groups, and thus is no effect marked by ``-''. }
    \begin{tabular}{l|cc|cc|cc}
        \toprule
        \multirow{2}{*}{Dataset} & \multicolumn{2}{c|}{Movielens}  & \multicolumn{2}{c|}{Amazon} & \multicolumn{2}{c}{Alipay} \\
        & AUUC & QINI & AUUC & QINI & AUUC & QINI \\
        \midrule
        CRec & \multicolumn{2}{c|}{-} & \multicolumn{2}{c|}{-} & \multicolumn{2}{c}{-} \\
                ORec &0.0337 & 0.0322 & 0.0025& 0.0043& 0.024 & 0.010 \\
        CRRec & 0.0338& 0.0322 &0.0065 & 0.0111& 0.037 & 0.033 \\
        \bottomrule
    \end{tabular}
    \label{tab:qini}
\end{table}

\subsubsection{RQ1} To show the promise of dataset construction via causal inference and the performance of the Meta-Controller-based recommendation, we conduct a range of experiments on Movielens, Amazon and Alipay. Specifically, we shows the uplift between different recommendation mechanisms in Figure~\ref{fig:auuc} and summarize the AUUC performance and their corresponding QINI scores compared to the plain session recommendation in Table~\ref{tab:qini}. The CTR performance of all recommendation mechanisms measured by AUC, HitRate and NDCG is summarized in Table~\ref{tab:ctr}. 

According to the uplift curve in Figure~\ref{fig:auuc}, both the on-device recommendation and the cloud-based re-fresh achieve the positive gain on the basis of the plain session recommendation. This demonstrates that the real-time feedback from the user is important to improve the recommendation performance, no matter it is incorporated by the lightweight on-device model or the cloud-based model. However, the comparison between the on-device recommendation and the cloud-based re-fresh are indefinite as shown in the right panel of Figure~\ref{fig:auuc}. In comparison with the plain session recommendation, we can find that the AUUC score and the QINI score of ORec and CRRec in Table~\ref{tab:qini} are positive, which further confirms that two treatments practically enhances the long-term gain of the session recommendation. Therefore, meta controller targets to distinguish their diverse merits and invoke the proper recommender. 

Regarding the CTR performance in Table~\ref{tab:ctr}, CRec is usually not better than ORec and CRRec, and MCRec generally achieves a larger improvement by properly mixing three type of mechanisms. In comparison, RMixRec that randomly mixes them only achieves the better performance than CRec on Movienlens and Amazon, and shows the minor improvement over ORec on Alipay. These results confirms CRec, ORec and CRRec are complementary, requiring to selectively activate one of recommendation mechanisms instead of the random choice.  In a nutshell, MCRec is an effective way to manage the device-cloud collaborative recommendation.

\begin{figure}[!t]
  \centering
  \includegraphics[width=0.9\linewidth]{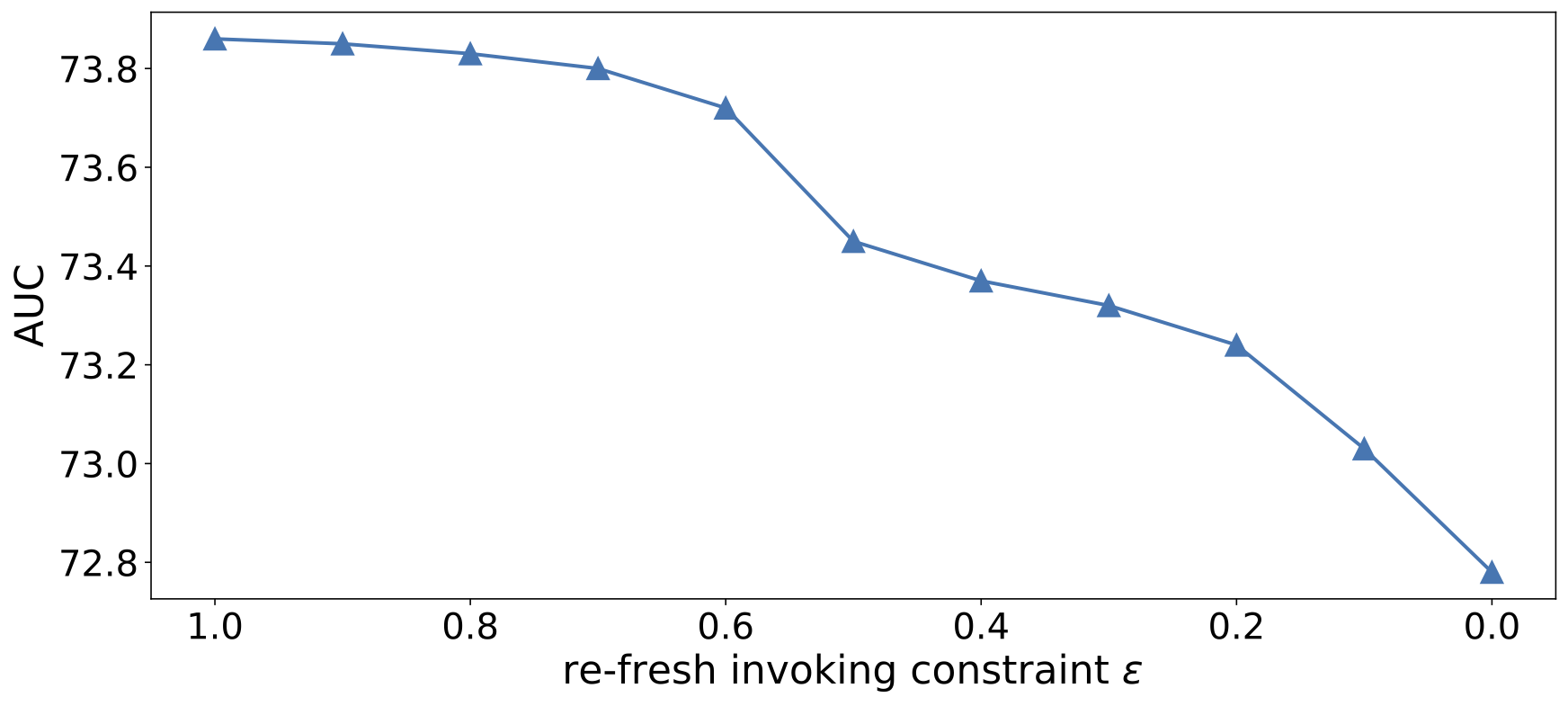}
  \caption{The trade-off between the model performance and the invoking budget $\epsilon$ of the re-fresh on the Alipay dataset.}\label{fig:trade-off}
\end{figure}

\begin{figure*}[!t]
  \centering
  \includegraphics[width=0.83\linewidth]{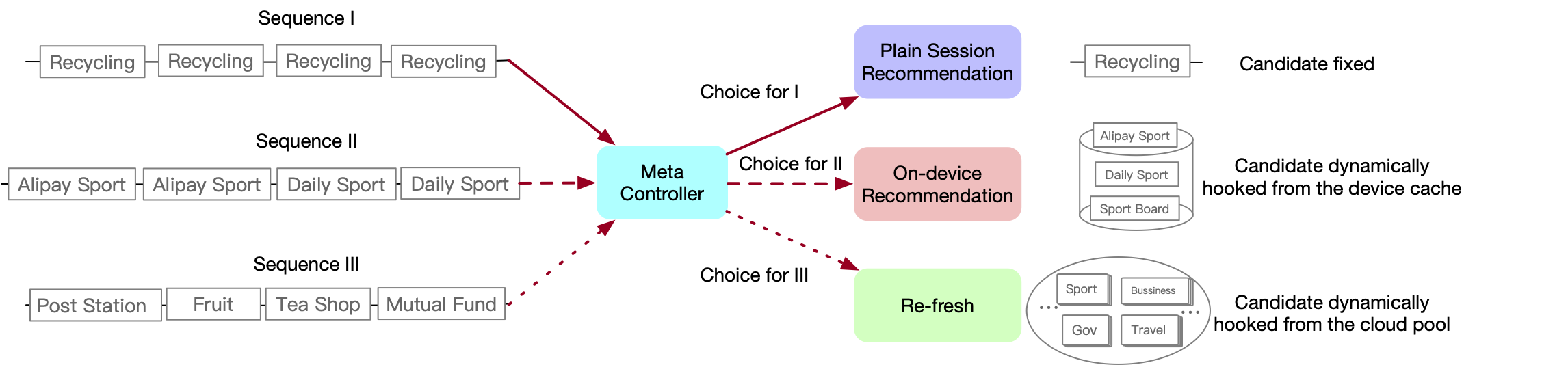}
  \caption{The exemplars from the Alipay background word dataset to characterize device-cloud collaborative recommendation via Meta Controller. Meta Controller captures the information from different interest-varying historical click sequences and makes decisions in invoking recommenders as respectively traced by the solid arrow, the dashed arrow and the dotted arrow.}\label{fig:exemplar}
\end{figure*}

\subsubsection{RQ2} To consider the computational constraint of the cloud resource, we apply the RSLS method defined by Eq.~\eqref{eq:rerm} to incorporate the invoking ratio $\epsilon$ on the cloud-based re-fresh. Specifically, we set different $\epsilon$ constraints \textit{e.g.,} $\epsilon=0.5$, $\epsilon=0.3$, $\epsilon=0.1$, which requires the invoking of the cloud-based re-fresh among all selection from three mechanisms not to outperform the constraint. By applying LS to learn the meta controller, the partial results of the CTR performance are summarized in Table~\ref{tab:ctr} and the trade-off between the recommendation performance of meta controller and $\epsilon$ is plot in Figure~\ref{fig:trade-off}. According to the results, we can see that with $\epsilon$ decreasing, namely increasing the invoking constraint about the cloud-based re-fresh, the performance of MCRec all approximately monotonically decreases. On one hand, this indicates the cloud-based re-fresh is an important way to complement the current session recommendation. On the other hand, it is also limited to the real-world computing resources available especially for the large-scale industrial scenarios where there are billions of customers. Therefore, it is still a trade-off between the better performance and more computation resources. Nevertheless, the discounted performance of meta controller (\textit{e.g.,} MCRec ($\epsilon=0.1$) in Table~\ref{tab:ctr}) still shows the gain by the proper mixing compared to CRec and ORec.

\begin{table}[!t]
    \centering
    \caption{The collaborative recommendation performance (HitRate$@$1 and AUC) of arbitrary two recommendation mechanisms on three datasets controlled by Meta Controller. \textcircled{1}, \textcircled{2} and \textcircled{3} respectively denotes CRec, ORec, CRRec. }
    \begin{tabular}{l|cc|cc|cc}
        \toprule
        \multirow{2}{*}{Dataset} & \multicolumn{2}{c|}{Movielens}  & \multicolumn{2}{c|}{Amazon} & \multicolumn{2}{c}{Alipay} \\
        & HR$@$1 & AUC & HR$@$1 & AUC & HR$@$1 & AUC \\
        \midrule
        \textcircled{1}+\textcircled{2} &36.72 &93.60 & 17.31& 77.95& 17.82 & 72.78 \\
        \textcircled{1}+\textcircled{3} &37.27 &93.88& 17.56& 78.98& 17.95 & 73.21 \\
        \textcircled{2}+\textcircled{3} &37.88&93.90 & 17.70& 79.05& 17.98 & 73.34 \\
        \bottomrule
    \end{tabular}
    \label{tab:twomix}
\end{table}

\subsubsection{RQ3} 
Here, we present more analysis about meta controller.

For the first ablation study, we discuss the degeneration of meta controller from the mixture of three kinds of recommendation mechanisms to arbitrary two types of recommendation mechanisms. Correspondingly, the arbitrary two of three sets during the sample construction stage are used and the 3-category classification problem is then reduced to 2-category classification in meta controller. As shown in Table~\ref{tab:twomix}, among three cases of collaboration, the combination between on-device recommender and the cloud-based re-fresh achieves the best performance. This indicates the merits of real-time features are important to improve the final model performance. However, in terms of the independent CRRec in Table~\ref{tab:ctr}, such 
performance is comparable on Alipay. One potential explanation is, in addition to the benefit from the real-time user behavior, the cloud-based recommender provides the sufficient capacity to consider the case of the slowly-changing user interests, which is more complementary to ORec and CRRec together.

\begin{figure}[!t]
  \centering
  \includegraphics[width=0.9\linewidth]{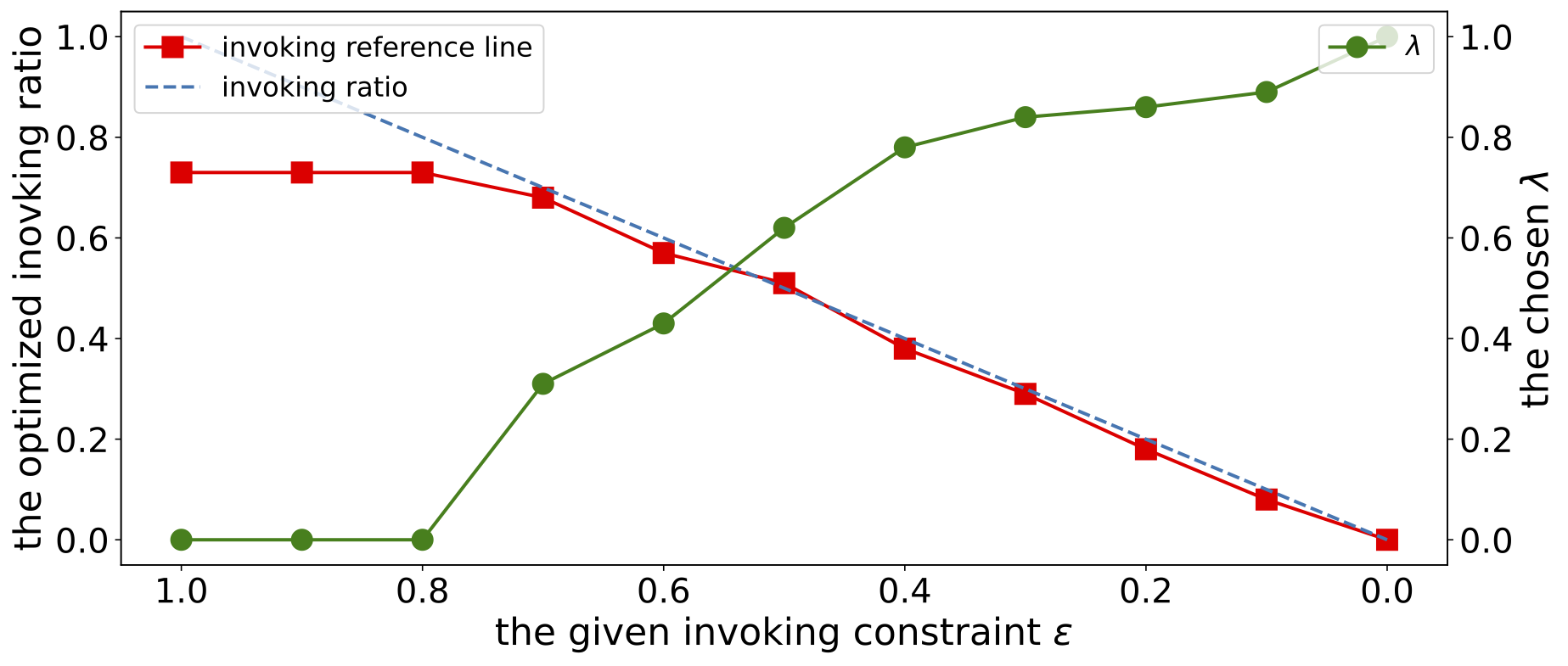}
  \caption{The $\lambda$ in Eq.~\ref{eq:rerm} chosen for different $\epsilon$ and the resulted invoking ratios (red line) on the Alipay dataset.}\label{fig:parameter}
\end{figure}

For the second ablation study, we investigate the effect of the hyperparameter $\lambda$ to control the invoking ratio under different constraints $\epsilon$. Considering the limited space, we only conduct the experiments on Alipay and illustrate the curves in Figure~\ref{fig:parameter}. From the figure, we can see that a larger $\lambda$ that introduces a stronger penalty on invoking the cloud-based re-fresh in Eq.~\eqref{eq:rerm}, leads to a lower invoking rate. The resulted invoking ratio (the read line) in meta controller almost satisfy the constraints characterized by the dashed line in Figure~\ref{fig:parameter}. Note that, when $\epsilon=1.0,~0.9,~0.8$, $\lambda$ does not change. It is because the original invoking ratio in meta controller is smaller than the constraint, and thus there is no need to penalize. In summary, our label smoothing method can flexibly handle the practical invoking constraints for the cloud-based re-fresh. 

To visualize how the meta controller manages the device-cloud collaboration, we present one exemplar from the Alipay dataset in Figure~\ref{fig:exemplar}. As shown in the figure, when the user behavior is lack of diversity, namely only focuses on one content, the plain session recommender is chosen by meta controller and recommend the fixed candidate. When the user behavior exhibits the small variance, namely interested in the homogeneous contents, the on-device recommender is chosen to slightly adjust the recommendation on the basis of the limited device cache. When the user interests dramatically change as reflected by the diverse historical behaviors, the meta controller invokes the cloud-based re-fresh to adjust the recommendation results from the large-scale cloud item pool. From these exemplars, meta controller dynamically 
senses the diversity of user interests and actives the corresponding recommender.

\section{Conclusion}
In this paper, we explore a novel device-cloud collaborative recommendation via meta controller. Different from on-device recommendation or the cloud-based recommendation, our method targets to manage the collaboration among these recommendation mechanisms by complementing their pros. To overcome the dataset absence issue, we propose an efficient offline sample construction method, which leverages the causal effect estimation technique to infer the counterfactual label. Besides, considering the real-world computational constraints, we introduces a label smoothing method to regularize the training procedure of meta controller. We demonstrate the promise of our proposed method on two public datasets and one large-scale real-world datasets. In the future, more recommendation scenarios could be explored from the perspective device-cloud collaborative recommendation to improve the efficiency of the conventional cloud-based recommendation.

\bibliographystyle{ACM-Reference-Format}
\bibliography{ref}


\newpage 
\appendix
\section{Reproducibility Details}
\subsection{Dataset Generation}
\textbf{Amazon}
and \textbf{Movielens-1M}
We preprocess the datasets to guarantee that users and items have been interacted at least 10 times in Amazon dataset, 20 times in Movielens-1M dataset. The rated movies and reviewed goods are regarded as positive samples. For both datasets, the last 5 interacted items of each user are kept for test and the left sequences are used for training. We randomly sample un-interacted movies or goods of the user as negative samples and the ratio between positive and negative samples is set to 1:4 in training sets and 1:100 in testing sets. To simulate the on-device recommendation process, we make different restrictions for cloud models and device models. Specifically, the cloud models take user sequences with a length up to 50 as inputs, but are not accessible to the user activities in the next session, where users make 5 further interactions. The device models, however, can make use of the most recent interactions in the current session, but are not accessible to the more historical behaviors saved in cloud. To train the CATE estimation model, we evaluate cloud models and device models on all samples and set treatment effect to 1 if the model could rank the positive sample to the first position.

\textbf{Alipay} dataset is collected from the Alipay App in the continuous 7 days. The treatment datasets are respectively collected under different treatment interventions online. Except training datasets aforementioned in the experimental parts, the test datasets consist of two parts, one part is for the CATE estimation and the other one is for the CTR comparison. 

\subsection{Parameter Settings}
Table~\ref{tab:params} shows the hyper-parameter setting on three datasets.

\begin{table}[t!]
\centering
\caption{Parameters and training schedules of meta controller and baselines on Amazon, Movielens, Alipay datasets}
\label{tab:params}
\begin{tabular}{@{}lll@{}}
\toprule
Dataset                     & Parameters                                     & Setting         \\ 
\midrule
\multirow{11}{*}{Amazon} &  cloud item embedding dimension                    & 32               \\ 
\cmidrule(l){2-3} 
                            & device item embedding dimension                    & 8               \\ \cmidrule(l){2-3} 
                            & cloud position embedding dimension                   & 32               \\ \cmidrule(l){2-3} 
                            & device position embedding dimension                   & 8               \\ \cmidrule(l){2-3} 
                            & encoder layer                                  & 2               \\ \cmidrule(l){2-3} 
                            & encoder size                                   & 32              \\ \cmidrule(l){2-3} 
                            & attention dimension(Q,K,V)                     & 32              \\ \cmidrule(l){2-3} 
                            & classifer dimension                            & 128,64          \\ \cmidrule(l){2-3} 
                            & activation function                            & tanh            \\ \cmidrule(l){2-3} 
                            & batch size                                     & 256             \\ \cmidrule(l){2-3} 
                            & learning rate                                  & 1e-3            \\ \cmidrule(l){2-3} 
                            & optimizer                                      & Adam            
                            \\ \midrule
\multirow{12}{*}{Movielens-1M} & item embedding dimension                    & 32               \\
\cmidrule(l){2-3} 
    & device item embedding dimension                    & 8               \\
\cmidrule(l){2-3} 
                            & cloud position embedding dimension                   & 32               \\ \cmidrule(l){2-3} 
                            & device position embedding dimension                   & 8               \\ \cmidrule(l){2-3} 
                            & encoder layer                                  & 2               \\ \cmidrule(l){2-3} 
                            & encoder size                                   & 32              \\ \cmidrule(l){2-3} 
                            & attention dimension(Q,K,V)                     & 32              \\ \cmidrule(l){2-3} 
                            & classifer dimension                            & 128,64          \\ \cmidrule(l){2-3} 
                            & activation function                            & tanh            \\ \cmidrule(l){2-3} 
                            & batch size                                     & 256             \\ \cmidrule(l){2-3} 
                            & learning rate                                  & 1e-3            \\ \cmidrule(l){2-3} 
                            & optimizer                                      & Adam            \\
                            \midrule
\multirow{14}{*}{Alipay}    & item embedding ($f_\cloud(\cdot)$ and all $g(\cdot)$) & 128 \\ \cmidrule(l){2-3} 
                            & item embedding ($f_\device(\cdot)$)            & 64               \\ \cmidrule(l){2-3} 
                            & sequence length                            & 50         \\
                            \cmidrule(l){2-3}
                            & learning rate                                  & 5e-3            \\ \cmidrule(l){2-3} 
                            & optimizer                                      & Adam            \\ \cmidrule(l){2-3} 
                            & training epoches                               & 10            \\ \cmidrule(l){2-3} 
                            & batch size                                     & 1024            \\ \cmidrule(l){2-3} 
                            & activation function                            & relu            \\ \cmidrule(l){2-3} 
                            & attention dimension(Q,K,V)                     & 128              \\ \cmidrule(l){2-3} 
                            & classifer dimension                            & 128-64-2         \\
                            \cmidrule(l){2-3}
                            & sequence length                            & 50         \\
                            \cmidrule(l){2-3}
                            & $\lambda$ &  see 4.2.3            \\ \bottomrule
\end{tabular}
\end{table}


\end{document}